\documentclass[11pt]{article}

\usepackage[preprint]{acl}

\usepackage{times}
\usepackage{latexsym}

\usepackage[T1]{fontenc}

\usepackage[utf8]{inputenc}

\usepackage{microtype}

\usepackage{inconsolata}

\usepackage{graphicx}
\usepackage{multirow}
\usepackage{amssymb}
\usepackage{amsmath}
\usepackage{booktabs}
\usepackage{pifont}

\usepackage[edges]{forest}
\usepackage{xcolor}
\usetikzlibrary{patterns}
\usepackage{rotating}
\usepackage{tikz}
\usetikzlibrary{positioning, shapes.geometric, arrows.meta, calc, shadows.blur}
\usepackage{enumitem}

\definecolor{tempMain}{HTML}{1565C0}
\definecolor{tempEM}{HTML}{0D47A1}
\definecolor{tempDD}{HTML}{1976D2}
\definecolor{tempFC}{HTML}{42A5F5}
\definecolor{coocMain}{HTML}{E65100}
\definecolor{coocOA}{HTML}{E65100}
\definecolor{coocSE}{HTML}{FF9800}
\definecolor{aggrMain}{HTML}{2E7D32}
\definecolor{aggrCA}{HTML}{1B5E20}
\definecolor{aggrSA}{HTML}{43A047}
\definecolor{audioMain}{HTML}{7B1FA2}
\definecolor{audioAA}{HTML}{6A1B9A}
\definecolor{audioEI}{HTML}{AB47BC}

\definecolor{taskMC}{HTML}{0D47A1}
\definecolor{taskBin}{HTML}{42A5F5}
\definecolor{taskOpen}{HTML}{6A1B9A}
\definecolor{taskCap}{HTML}{AB47BC}
\definecolor{lenS}{HTML}{43A047}
\definecolor{lenM}{HTML}{FF9800}
\definecolor{lenL}{HTML}{E65100}
\definecolor{lenSt}{HTML}{C62828}

\definecolor{checkgreen}{HTML}{2E7D32}
\definecolor{crossred}{HTML}{C62828}

\newcommand{\cmark}{\textcolor{checkgreen}{\ding{51}}}
\newcommand{\xmark}{\textcolor{crossred}{\ding{55}}}

\title{Distorted or Fabricated? A Survey on Hallucination in Video LLMs}

\author{
 \textbf{Yiyang Huang\textsuperscript{1}},
 \textbf{Yitian Zhang\textsuperscript{1}},
 \textbf{Yizhou Wang\textsuperscript{1}},
 \textbf{Mingyuan Zhang\textsuperscript{1}}
\\
 \textbf{Liang Shi\textsuperscript{1}},
 \textbf{Huimin Zeng\textsuperscript{1}},
 \textbf{Yun Fu\textsuperscript{1,2}}
\\
  \textsuperscript{1}Department of Electrical and Computer Engineering, Northeastern University\\
 \textsuperscript{2}Khoury College of Computer Science, Northeastern University
\\
 \small{
   \textbf{Correspondence:} \href{mailto:huang.yiyan@northeastern.edu}{huang.yiyan@northeastern.edu},
   \href{mailto:y.fu@northeastern.edu}{y.fu@northeastern.edu}
 }
\\
 \small{
    \textbf{Continuously updated curated list:}
    \href{https://github.com/hukcc/Awesome-Video-Hallucination}{https://github.com/hukcc/Awesome-Video-Hallucination}
    }
}

\begin{document}
\maketitle

\begin{abstract}
Despite significant progress in video-language modeling, hallucinations remain a persistent challenge in Video Large Language Models (Vid-LLMs), referring to outputs that appear plausible yet contradict the content of the input video. This survey presents a comprehensive analysis of hallucinations in Vid-LLMs and introduces a systematic taxonomy that categorizes them into two core types: dynamic distortion and content fabrication, each comprising two subtypes with representative cases. Building on this taxonomy, we review recent advances in the evaluation and mitigation of hallucinations, covering key benchmarks, metrics, and intervention strategies. We further analyze the root causes of dynamic distortion and content fabrication, which often result from limited capacity for temporal representation and insufficient visual grounding. These insights inform several promising directions for future work, including the development of motion-aware visual encoders and the integration of counterfactual learning techniques. This survey consolidates scattered progress to foster a systematic understanding of hallucinations in Vid-LLMs, laying the groundwork for building robust and reliable video-language systems. 
\end{abstract}

\section{Introduction}
\label{sec:introduction}
Video Large Language Models (Vid-LLMs) extend the capabilities of vision-language systems from static images to temporally coherent video inputs, enabling tasks such as action recognition, temporal reasoning, and audio-visual understanding~\cite{VideoLLaMA2023,VideoChatGPT2023,VideoChat2023,VideoLLaVA2023,InternVideo22024,VITA}. Despite recent advances, these models remain susceptible to hallucinations, producing outputs that appear plausible and coherent yet contradict the actual content of the video. This issue poses reliability and safety risks in safety-critical domains, including embodied AI~\cite{Embodied} and autonomous driving~\cite{Driving}.

\begin{figure*}[t]
\centering
\includegraphics[width=1.0\linewidth]{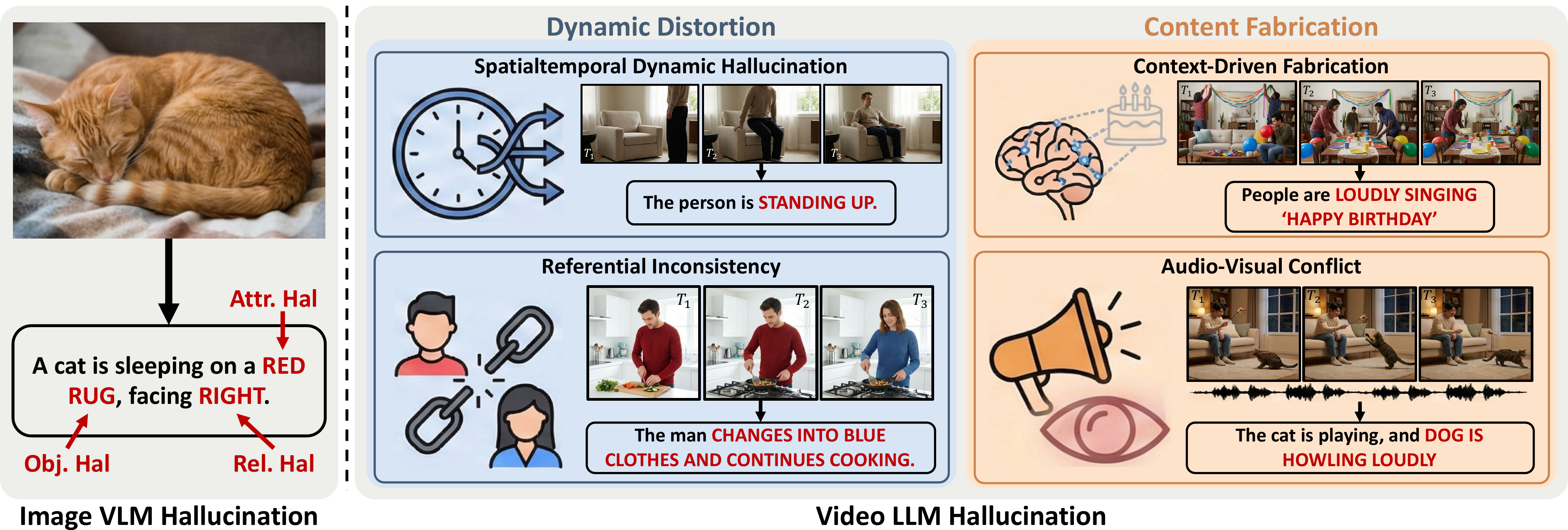}
\caption{Taxonomy of hallucinations in Vid-LLMs. Video models exhibit unique hallucination types beyond static errors in images. These fall into two main categories: (1) Dynamic Distortion, which includes spatiotemporal misrepresentation and referential inconsistency; (2) Content Fabrication, which includes hallucinations influenced by statistical priors and cases where auditory input overrides visual evidence.}
\label{fig:hallucination_illustration}
\end{figure*}

While hallucinations have been extensively surveyed in image-based vision-language models (VLMs)~\cite{imageVLMSurvey1,imageVLMSurvey2}, the inherent complexity of video's temporal structure, motion dynamics, and audio-visual integration complicates the direct application of these insights to the video domain. To address this gap, this survey presents a video-specific, mechanism-driven taxonomy that classifies hallucinations into two primary types: dynamic distortion, where the model misrepresents the spatiotemporal evolution or referential consistency of entities and scenes; and content fabrication, where outputs are influenced by prior knowledge or dominated by audio modality.

Building on this taxonomy, we review recent advances in the evaluation and mitigation of hallucinations in Vid-LLMs, with a focus on key benchmarks, metrics, and intervention strategies. 
Dynamic distortion includes hallucinations in spatiotemporal dynamics, such as incorrect event ordering~\cite{VidHalluc,SEASON,SmartSight}, inaccurate duration estimation~\cite{VideoHallucer,OVBench,TemporalInsightEnhancement}, and frequency miscounting~\cite{HAVEN,VidHal}, as well as referential inconsistency, where the model conflates different characters~\cite{EGOILLUSION,MESH} or scenes~\cite{ELV-Halluc,Vista-llama,VideoPLR}. 
Content fabrication includes context-driven hallucinations, where commonly co-occurring object–action~\cite{SANTA,VideoHallu} or scene–event~\cite{MASH-VLM,EventHallusion,PaMi-VDPO} patterns lead to unsupported inferences; and audio-visual conflict, where dominant auditory cues override visual evidence, resulting in hallucinated actions~\cite{AVHBench,AVCD} or emotional states~\cite{EmotionHallucer}.

Further analysis reveals the underlying mechanisms of dynamic distortion and content fabrication. Dynamic distortion often results from missing fine-grained motion cues due to limited temporal encoding~\cite{VideoPerceiver,TempCompass}, and is further exacerbated in long videos by weak long-range memory~\cite{MASH-VLM} and poor temporal localization~\cite{SEASON}. In contrast, content fabrication arises from insufficient visual grounding~\cite{NOAH}, allowing pretrained priors~\cite{VideoHallu} or dominant audio signals~\cite{CMM} to override visual evidence. 

In light of these underlying mechanisms, promising research directions include developing motion-aware architectures\cite{MASS} that retain fine-grained temporal features to strengthen the alignment between visual perception and temporal reasoning. In addition, counterfactual training strategies that disentangle visual evidence from prior knowledge\cite{Taming} offer a principled approach to mitigating content fabrication by encouraging models to ground predictions more faithfully in the visual input.

\textbf{Comparison with existing surveys.}
Hallucination has been extensively studied in LLMs and image-based VLMs~\cite{LLMSurvey1,LLMSurvey2,imageVLMSurvey1,imageVLMSurvey2}. While MLLM hallucination surveys~\cite{MLLMSurvey1, MLLMSurvey2} include video alongside other modalities, their discussion of video hallucination remains superficial, offering only brief mentions of benchmarks and mitigation strategies without structural or causal analysis. In contrast, this survey presents the first mechanism-driven taxonomy of hallucinations in Vid-LLMs. We propose a layered classification framework (Fig.~\ref{fig:taxonomy}), conduct a broader and more detailed review of existing literature, and analyze the underlying causes of hallucinations. Building on this analysis, we outline future directions that align closely with identified causes, benchmark coverage, and mitigation strategies, offering a cohesive roadmap toward hallucination-resilient Vid-LLMs.

\section{Definition and Scope}
\label{sec:definition}

\textbf{Definition.}  
We define \textit{video hallucination} as cases where a Vid-LLM generates textual outputs that are linguistically coherent and contextually plausible, yet contradict the observable spatiotemporal evidence in the input video.

\textbf{Distinction from Static Image Hallucination.} 
While hallucinations in image-based VLMs, such as those involving objects, attributes, and relations, have been well studied~\cite{imageVLMSurvey1, imageVLMSurvey2}, the video modality introduces a temporal dimension that fundamentally alters the problem. Unlike static inputs, videos require reasoning over causality, temporal grounding, motion dynamics, and audio-visual integration. These aspects are beyond the scope of traditional static metrics. Our taxonomy explicitly captures these temporal and multimodal challenges, distinguishing video hallucination from image-based settings.

\section{Taxonomy of Video Hallucinations}
\label{sec:taxonomy}

\definecolor{tempMain}{HTML}{1565C0}
\definecolor{tempEM}{HTML}{0D47A1}
\definecolor{tempDD}{HTML}{1976D2}
\definecolor{tempFC}{HTML}{42A5F5}

\definecolor{coocMain}{HTML}{E65100}
\definecolor{coocOA}{HTML}{E65100}
\definecolor{coocSE}{HTML}{FF9800}

\definecolor{aggrMain}{HTML}{2E7D32}
\definecolor{aggrCA}{HTML}{1B5E20}
\definecolor{aggrSA}{HTML}{43A047}

\definecolor{audioMain}{HTML}{7B1FA2}
\definecolor{audioAA}{HTML}{6A1B9A}
\definecolor{audioEI}{HTML}{AB47BC}

\definecolor{metaDistort}{HTML}{00838F}
\definecolor{metaFabric}{HTML}{C2185B}

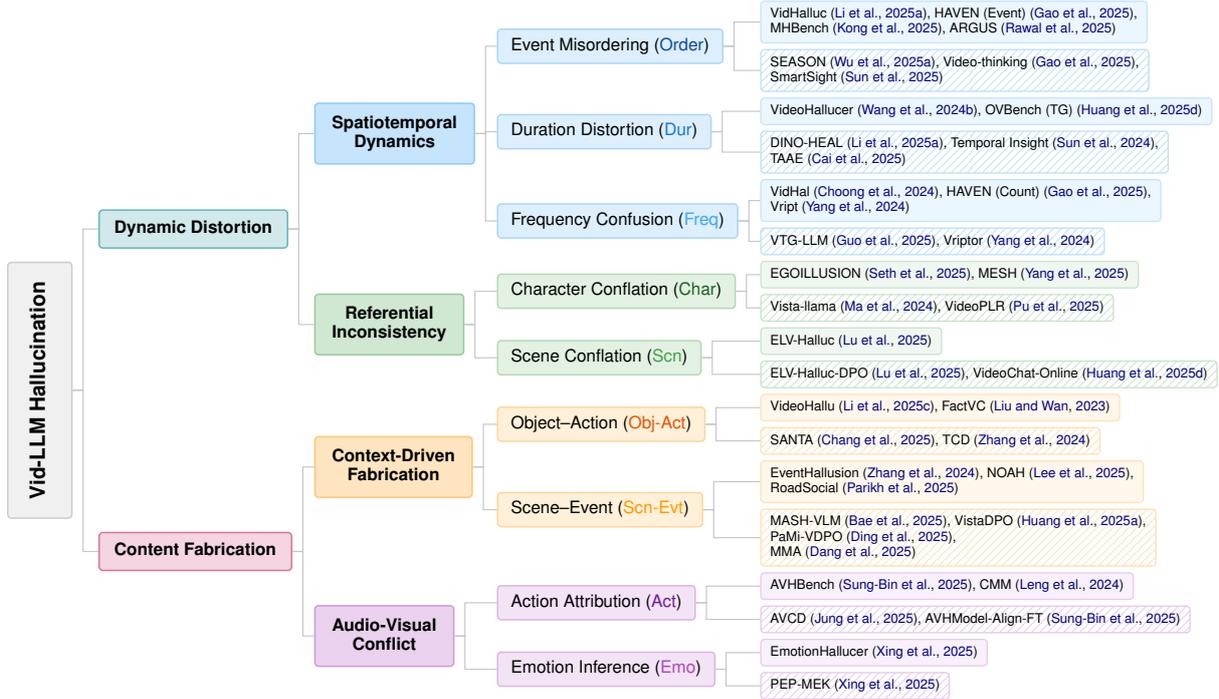
\begin{figure*}[htbp]
\centering
\resizebox{\textwidth}{!}{%
\begin{forest}
  for tree={
    grow'=0,
    rounded corners=2pt,
    line width=0.8pt,
    align=center,
    inner xsep=5pt,
    inner ysep=3pt,
    l sep=4mm,
    s sep=1mm,
    edge={draw=gray!45, line width=0.6pt},
    forked edge,
    font=\small\sffamily,
    text=black,
    anchor=west,
    child anchor=west,
    parent anchor=east,
  },
  root/.style={fill=gray!12, draw=gray!40, font=\bfseries\normalsize\sffamily, align=center, inner xsep=10pt, inner ysep=10pt, rotate=90, anchor=center, tier=root},
  metaD/.style={fill=metaDistort!18, draw=metaDistort!60, font=\bfseries\small\sffamily, align=center, tier=meta, inner xsep=8pt, inner ysep=5pt, text=black},
  metaF/.style={fill=metaFabric!18, draw=metaFabric!60, font=\bfseries\small\sffamily, align=center, tier=meta, inner xsep=8pt, inner ysep=5pt, text=black},
  L1T/.style={fill=tempFC!30, draw=tempFC!60, font=\bfseries\small\sffamily, align=center, tier=L1, inner xsep=9pt, inner ysep=6pt, text=black},
  L1C/.style={fill=coocSE!25, draw=coocSE!55, font=\bfseries\small\sffamily, align=center, tier=L1, inner xsep=9pt, inner ysep=6pt, text=black},
  L1A/.style={fill=aggrSA!25, draw=aggrSA!55, font=\bfseries\small\sffamily, align=center, tier=L1, inner xsep=9pt, inner ysep=6pt, text=black},
  L1V/.style={fill=audioEI!25, draw=audioEI!55, font=\bfseries\small\sffamily, align=center, tier=L1, inner xsep=9pt, inner ysep=6pt, text=black},
  L2T/.style={fill=tempFC!18, draw=tempFC!45, font=\small\sffamily, align=center, tier=L2, inner xsep=7pt, inner ysep=4pt},
  L2C/.style={fill=coocSE!15, draw=coocSE!40, font=\small\sffamily, align=center, tier=L2, inner xsep=7pt, inner ysep=4pt},
  L2A/.style={fill=aggrSA!15, draw=aggrSA!40, font=\small\sffamily, align=center, tier=L2, inner xsep=7pt, inner ysep=4pt},
  L2V/.style={fill=audioEI!15, draw=audioEI!40, font=\small\sffamily, align=center, tier=L2, inner xsep=7pt, inner ysep=4pt},
  benchT/.style={fill=tempFC!10, draw=tempFC!30, line width=0.7pt, font=\scriptsize\sffamily, align=left, tier=L3, inner xsep=5pt, inner ysep=3pt},
  benchC/.style={fill=coocSE!8, draw=coocSE!28, line width=0.7pt, font=\scriptsize\sffamily, align=left, tier=L3, inner xsep=5pt, inner ysep=3pt},
  benchA/.style={fill=aggrSA!8, draw=aggrSA!28, line width=0.7pt, font=\scriptsize\sffamily, align=left, tier=L3, inner xsep=5pt, inner ysep=3pt},
  benchV/.style={fill=audioEI!8, draw=audioEI!28, line width=0.7pt, font=\scriptsize\sffamily, align=left, tier=L3, inner xsep=5pt, inner ysep=3pt},
  methT/.style={pattern=north east lines, pattern color=tempFC!30, draw=tempFC!35, line width=0.7pt, font=\scriptsize\sffamily, align=left, tier=L3, inner xsep=5pt, inner ysep=3pt},
  methC/.style={pattern=north east lines, pattern color=coocSE!25, draw=coocSE!30, line width=0.7pt, font=\scriptsize\sffamily, align=left, tier=L3, inner xsep=5pt, inner ysep=3pt},
  methA/.style={pattern=north east lines, pattern color=aggrSA!25, draw=aggrSA!30, line width=0.7pt, font=\scriptsize\sffamily, align=left, tier=L3, inner xsep=5pt, inner ysep=3pt},
  methV/.style={pattern=north east lines, pattern color=audioEI!25, draw=audioEI!30, line width=0.7pt, font=\scriptsize\sffamily, align=left, tier=L3, inner xsep=5pt, inner ysep=3pt},
  [{Vid-LLM Hallucination}, root
    [{\textbf{Dynamic Distortion}}, metaD
      [{\textbf{Spatiotemporal}\\{\textbf{Dynamics}}}, L1T
      [{Event Misordering ({\footnotesize\textcolor{tempEM}{Order}})}, L2T
        [{VidHalluc~\cite{VidHalluc}, HAVEN (Event)~\cite{HAVEN},\\MHBench~\cite{MHBench}, ARGUS~\cite{ARGUS}}, benchT]
        [{SEASON~\cite{SEASON}, Video-thinking~\cite{HAVEN},\\SmartSight~\cite{SmartSight}}, methT]
      ]
      [{Duration Distortion ({\footnotesize\textcolor{tempDD}{Dur}})}, L2T
        [{VideoHallucer~\cite{VideoHallucer}, OVBench (TG)~\cite{OVBench}}, benchT]
        [{DINO-HEAL~\cite{VidHalluc}, Temporal Insight~\cite{TemporalInsightEnhancement},\\TAAE~\cite{TAAE}}, methT]
      ]
      [{Frequency Confusion ({\footnotesize\textcolor{tempFC}{Freq}})}, L2T
        [{VidHal~\cite{VidHal}, HAVEN (Count)~\cite{HAVEN},\\Vript~\cite{Vript}}, benchT]
        [{VTG-LLM~\cite{VTG-LLM}, Vriptor~\cite{Vript}}, methT]
      ]
    ]
      [{\textbf{Referential}\\{\textbf{Inconsistency}}}, L1A
        [{Character Conflation ({\footnotesize\textcolor{aggrCA}{Char}})}, L2A
          [{EGOILLUSION~\cite{EGOILLUSION}, MESH~\cite{MESH}}, benchA]
          [{Vista-llama~\cite{Vista-llama}, VideoPLR~\cite{VideoPLR}}, methA]
        ]
        [{Scene Conflation ({\footnotesize\textcolor{aggrSA}{Scn}})}, L2A
          [{ELV-Halluc~\cite{ELV-Halluc}}, benchA]
          [{ELV-Halluc-DPO~\cite{ELV-Halluc}, VideoChat-Online~\cite{OVBench}}, methA]
        ]
      ]
    ]
    [{\textbf{Content Fabrication}}, metaF
      [{\textbf{Context-Driven}\\{\textbf{Fabrication}}}, L1C
      [{Object--Action ({\footnotesize\textcolor{coocOA}{Obj-Act}})}, L2C
        [{VideoHallu~\cite{VideoHallu}, FactVC~\cite{FactVC}}, benchC]
        [{SANTA~\cite{SANTA}, TCD~\cite{EventHallusion}}, methC]
      ]
      [{Scene--Event ({\footnotesize\textcolor{coocSE}{Scn-Evt}})}, L2C
        [{EventHallusion~\cite{EventHallusion}, NOAH~\cite{NOAH},\\RoadSocial~\cite{RoadSocial}}, benchC]
        [{MASH-VLM~\cite{MASH-VLM}, VistaDPO~\cite{VistaDPO},\\PaMi-VDPO~\cite{PaMi-VDPO}, \\MMA~\cite{MMA}}, methC]
      ]
    ]
      [{\textbf{Audio-Visual}\\{\textbf{Conflict}}}, L1V
      [{Action Attribution ({\footnotesize\textcolor{audioAA}{Act}})}, L2V
        [{AVHBench~\cite{AVHBench}, CMM~\cite{CMM}}, benchV]
        [{AVCD~\cite{AVCD}, AVHModel-Align-FT~\cite{AVHBench}}, methV]
      ]
      [{Emotion Inference ({\footnotesize\textcolor{audioEI}{Emo}})}, L2V
        [{EmotionHallucer~\cite{EmotionHallucer}}, benchV]
        [{PEP-MEK~\cite{EmotionHallucer}}, methV]
        ]
      ]
    ]
  ]
\end{forest}
}
\caption{Mechanism-driven taxonomy of Vid-LLM hallucinations. \textcolor{metaDistort}{\textbf{Dynamic Distortion}}: entities are perceived but their spatiotemporal evolution or identity is misinterpreted, including \textcolor{tempFC}{Spatiotemporal Dynamics} (Order/Dur/Freq) and \textcolor{aggrSA}{Referential Inconsistency} (Char/Scn). \textcolor{metaFabric}{\textbf{Content Fabrication}}: outputs lack visual evidence and are driven by priors, including \textcolor{coocSE}{Context-Driven Fabrication} (Obj-Act/Scn-Evt) and \textcolor{audioEI}{Audio-Visual Conflict} (Act/Emo). Solid fill denotes benchmarks; striped fill indicates mitigation methods.}
\label{fig:taxonomy}
\end{figure*}

As discussed in Section~\ref{sec:definition}, the temporal and multimodal nature of video poses challenges beyond static image settings. To address these, we propose a mechanism-driven taxonomy of \textit{dynamic-level hallucinations} unique to video, categorizing them by visually observable failure modes rather than input attributes (e.g., audio, length, or genre), which we treat as \textit{conditioning factors} affecting hallucination severity. This design is motivated by the observation that similar failure modes (e.g., dynamic relation errors or prior-driven completions) arise across input settings; using input attributes as primary axes would therefore separate structurally identical failures and hinder cross-setting comparability. The taxonomy captures failures in temporal reasoning and cross-modal alignment, and classifies hallucinations into \textit{Dynamic Distortion} and \textit{Content Fabrication}, each with two subtypes and representative cases (Figure~\ref{fig:taxonomy}). It provides a unified basis for evaluation and targeted mitigation (Sections~\ref{sec:benchmarks} and \ref{sec:mitigation}).

\subsection{Dynamic Distortion}
This category refers to situations where the model correctly detects entities but misrepresents their temporal progression or referential consistency. It includes two subtypes: \textit{Spatiotemporal Dynamics}, involving errors in event ordering, duration, or frequency; and \textit{Referential Inconsistency}, where characters or scenes are conflated across temporal boundaries. 

\textbf{Spatiotemporal Dynamics.}  
These hallucinations arise when the model correctly identifies relevant events but fails to model their temporal relationships. Typical cases include event misordering, such as reversing action causality or misinterpreting motion direction and trajectory~\cite{VidHalluc,HAVEN,SEASON,SmartSight}; duration distortion, where the model over- or underestimates the length of an action~\cite{VideoHallucer,OVBench,TemporalInsightEnhancement}; and frequency confusion, in which repeated actions are miscounted~\cite{HAVEN,VidHal}.

\definecolor{notHal}{HTML}{9E9E9E}
\definecolor{isHal}{HTML}{212121}

\definecolor{tempMain}{HTML}{1565C0}
\definecolor{tempEM}{HTML}{0D47A1}
\definecolor{tempDD}{HTML}{1976D2}
\definecolor{tempFC}{HTML}{42A5F5}

\definecolor{coocMain}{HTML}{E65100}
\definecolor{coocOA}{HTML}{E65100}
\definecolor{coocSE}{HTML}{FF9800}

\definecolor{aggrMain}{HTML}{2E7D32}
\definecolor{aggrCA}{HTML}{1B5E20}
\definecolor{aggrSA}{HTML}{43A047}

\definecolor{audioMain}{HTML}{7B1FA2}
\definecolor{audioAA}{HTML}{6A1B9A}
\definecolor{audioEI}{HTML}{AB47BC}

\definecolor{metaDistort}{HTML}{00838F}
\definecolor{metaFabric}{HTML}{C2185B}

\begin{figure*}[htbp]
\centering
\resizebox{\textwidth}{!}{%
\begin{forest}
  for tree={
    grow'=0,
    rounded corners=2pt,
    line width=0.8pt,
    align=left,                     
    inner xsep=6pt,                 
    inner ysep=4pt,
    l sep=9mm,                      
    s sep=1.8mm,                    
    edge={draw=gray!40, line width=0.6pt},
    forked edge,
    font=\small\sffamily,
    text=black,
    anchor=west,
    child anchor=west,
    parent anchor=east,
  },
  root/.style={
    fill=gray!12, draw=gray!40,
    font=\bfseries\normalsize\sffamily,
    rotate=90, anchor=center,
    align=center,
  },
  L1_NH/.style={
    fill=notHal!8, draw=notHal!45,
    font=\bfseries\small\sffamily,
    align=left,
    inner xsep=4pt, inner ysep=4pt, 
    text width=28mm,                
    tier=L1,
  },
  L1_H/.style={
    fill=isHal!6, draw=isHal!70,
    font=\bfseries\small\sffamily,
    align=left,
    inner xsep=4pt, inner ysep=4pt, 
    text width=28mm,
    tier=L1,
  },
  metaD/.style={
    fill=metaDistort!14, draw=metaDistort!55,
    font=\bfseries\small\sffamily,
    align=left,
    inner xsep=4pt, inner ysep=4pt, 
    text width=55mm,                
    tier=meta,
  },
  metaF/.style={
    fill=metaFabric!14, draw=metaFabric!55,
    font=\bfseries\small\sffamily,
    align=left,
    inner xsep=4pt, inner ysep=4pt, 
    text width=55mm,
    tier=meta,
  },
  subT/.style={
    fill=tempFC!20, draw=tempFC!50,
    font=\bfseries\small\sffamily,
    align=left,
    inner xsep=6pt, inner ysep=4pt,
    text width=76mm,                
    tier=sub,
  },
  subA/.style={
    fill=aggrSA!20, draw=aggrSA!50,
    font=\bfseries\small\sffamily,
    align=left,
    inner xsep=6pt, inner ysep=4pt,
    text width=76mm,
    tier=sub,
  },
  subC/.style={
    fill=coocSE!20, draw=coocSE!50,
    font=\bfseries\small\sffamily,
    align=left,
    inner xsep=6pt, inner ysep=4pt,
    text width=76mm,
    tier=sub,
  },
  subV/.style={
    fill=audioEI!20, draw=audioEI!50,
    font=\bfseries\small\sffamily,
    align=left,
    inner xsep=6pt, inner ysep=4pt,
    text width=76mm,
    tier=sub,
  },
  leafT/.style={
    fill=tempFC!10, draw=tempDD!35,
    font=\scriptsize\sffamily, align=left,
    line width=0.5pt, inner xsep=4pt, inner ysep=3pt, 
    text width=52mm,                
    tier=leaf,
  },
  leafA/.style={
    fill=aggrSA!10, draw=aggrCA!35,
    font=\scriptsize\sffamily, align=left,
    line width=0.5pt, inner xsep=4pt, inner ysep=3pt,
    text width=52mm,
    tier=leaf,
  },
  leafC/.style={
    fill=coocSE!10, draw=coocOA!35,
    font=\scriptsize\sffamily, align=left,
    line width=0.5pt, inner xsep=4pt, inner ysep=3pt,
    text width=52mm,
    tier=leaf,
  },
  leafV/.style={
    fill=audioEI!10, draw=audioAA!35,
    font=\scriptsize\sffamily, align=left,
    line width=0.5pt, inner xsep=4pt, inner ysep=3pt,
    text width=52mm,
    tier=leaf,
  },
  leafNH/.style={
    fill=notHal!6, draw=notHal!40,
    font=\scriptsize\sffamily, align=left,
    line width=0.5pt, inner xsep=4pt, inner ysep=3pt,
    text width=52mm,
    tier=leaf,
  }
  [{Video Hallucination?}, root
    [{\textbf{No}\\[-1pt] \mdseries\scriptsize Faithful to video}, L1_NH
      [{\textcolor{notHal}{\textbf{Stop:}} Output is consistent with the video input.}, leafNH]
    ]
    [{\textbf{Yes}\\[-1pt] \mdseries\scriptsize Hallucination detected}, L1_H
      [{\textbf{Dynamic Distortion}\\ \mdseries\scriptsize Visual evidence exists; relations mis-modeled.}, metaD
        [{\textbf{Spatiotemporal Dynamics}\\ \mdseries\scriptsize Failure to model temporal relationships.}, subT
          [{\textcolor{tempDD}{\textbf{Order:}} Incorrect temporal order or direction}, leafT]
          [{\textcolor{tempDD}{\textbf{Dur:}} Incorrect duration or boundary}, leafT]
          [{\textcolor{tempDD}{\textbf{Freq:}} Incorrect count or repetition}, leafT]
        ]
        [{\textbf{Referential Inconsistency}\\ \mdseries\scriptsize Conflation of entities or scenes across segments.}, subA
          [{\textcolor{aggrCA}{\textbf{Char:}} Identity confusion}, leafA]
          [{\textcolor{aggrCA}{\textbf{Scn:}} Scene confusion}, leafA]
        ]
      ]
      [{\textbf{Content Fabrication}\\ \mdseries\scriptsize No clear visual evidence; prior- or audio-driven.}, metaF
        [{\textbf{Context-Driven Fabrication}\\ \mdseries\scriptsize Reliance on statistical associations rather than visual grounding.}, subC
          [{\textcolor{coocOA}{\textbf{Obj-Act:}} Object-based completion}, leafC]
          [{\textcolor{coocOA}{\textbf{Scn-Evt:}} Scene-triggered event completion}, leafC]
        ]
        [{\textbf{Audio-Visual Conflict}\\ \mdseries\scriptsize Audio dominance overriding or contradicting visual evidence.}, subV
          [{\textcolor{audioAA}{\textbf{Act:}} Action contradicts visuals}, leafV]
          [{\textcolor{audioAA}{\textbf{Emo:}} Emotion contradicts visuals}, leafV]
        ]
      ]
    ]
  ]
\end{forest}
}%
\caption{Decision checklist for Vid-LLM hallucinations. The \textcolor{isHal}{\textbf{Yes}} branch tracks hallucination detection, while \textcolor{notHal}{\textbf{No}} indicates faithful grounding. \textcolor{metaDistort}{\textbf{Dynamic Distortion}}: visual evidence exists but relations fail, including \textcolor{tempFC}{Spatiotemporal Dynamics} (wrong order, duration, or count) and \textcolor{aggrSA}{Referential Inconsistency} (identity or scene confusion). \textcolor{metaFabric}{\textbf{Content Fabrication}}: visuals are absent and outputs rely on priors, including \textcolor{coocSE}{Context-Driven Fabrication} (prior-driven completion) and \textcolor{audioEI}{Audio-Visual Conflict} (audio contradicts visuals).}
\label{fig:decision_checklist_tree}
\end{figure*}
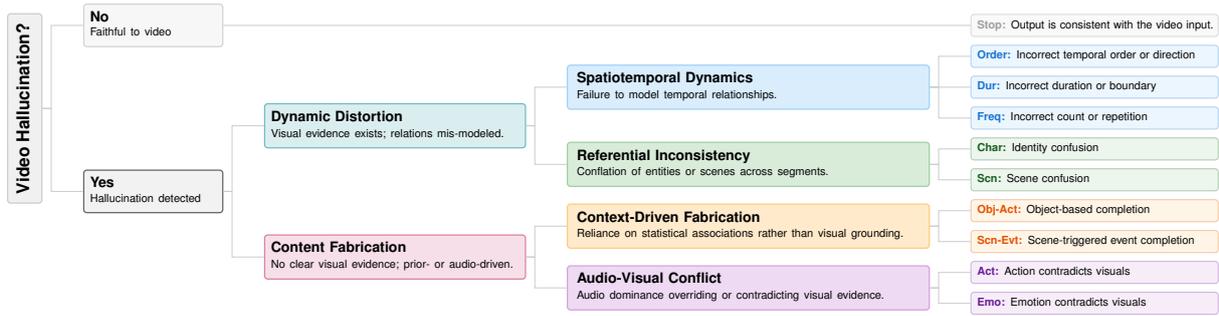

\textbf{Referential Inconsistency}  
These hallucinations refers to semantic-level failures where the model conflates distinct entities or scenes across temporal boundaries, producing blended descriptions that obscure segment distinctions. These errors arise when content from separate time spans is incorrectly merged into a single entity- or scene-level statement, even when visual cues could distinguish them. Such inconsistency typically appears in two forms: character conflation, where different individuals across scenes are mistakenly treated as the same person~\cite{EGOILLUSION,MESH}; and scene conflation, where actions or settings from distinct contexts are combined into a single narrative~\cite{ELV-Halluc,VideoPLR}.

\subsection{Content Fabrication}
This category covers cases where the model produces outputs that lack grounding in visual evidence and are instead influenced by learned priors. It includes \textit{context-driven fabrication}, where common object–action or scene–event associations result in unsupported predictions, and \textit{audio-visual conflict}, where auditory cues override visual input.

\textbf{Context-Driven Fabrication}  
This type of hallucination arises when the model relies on statistical associations from training data rather than grounding its predictions in visual evidence. An error is considered context-driven fabrication when the predicted action or event lacks visual support in the current observation window but is triggered by the presence of associated objects or scenes. It typically appears in two forms: object–action fabrication and scene–event fabrication. 
Object–action fabrication~\cite{SANTA,VideoHallu,FactVC} occurs when the presence of an object leads to incorrect action inference despite lacking motion cues. Scene–event fabrication~\cite{MASH-VLM,EventHallusion,PaMi-VDPO,MMA,VistaDPO} happens when typical events are predicted solely from background settings.

\textbf{Audio-Visual Conflict}  
This type of fabrication occurs when dominant or misleading audio cues override visual evidence, leading the model to generate outputs that align more with the audio than the video. Typical cases include hallucinated actions triggered by background sounds~\cite{AVHBench,AVCD}, and emotion inference based on vocal tone rather than facial expression~\cite{EmotionHallucer}.

\subsection{Taxonomy Separability and Exclusivity}
Our taxonomy is empirically separable and mutually exclusive, realized as a decision process based on visually observable evidence. Given a hallucinated output, the classification follows a single criterion: whether the claim is supported by visual evidence. If visual evidence exists but its spatiotemporal relations or cross-segment consistency are mis-modeled, the error is classified as \textit{Dynamic Distortion}; if no clear visual evidence is present and the claim is instead driven by statistical priors or non-visual cues, it is classified as \textit{Content Fabrication}.
For example, consider the boundary case where a model outputs ``The person is STANDING UP.'' If a sit--stand transition is visible but misrepresented (e.g., reversed), the error is \textit{Dynamic Distortion}; if no such transition is visible, it is \textit{Content Fabrication}. Similarly, Audio-Visual Conflict is categorized as a subtype of \textit{Content Fabrication}, as it reflects audio-dominant reasoning that contradicts or exceeds visual evidence. This decision process is summarized as a checklist (Figure~\ref{fig:decision_checklist_tree}), ensuring consistent and non-overlapping categorization across settings.

\begin{table*}[htbp]
\centering
\caption{Summary of video hallucination benchmarks. Video \textbf{length} and \textbf{domain} act as conditioning factors affecting hallucination severity: longer videos exacerbate referential inconsistency and long-range dynamic distortion, while domain-specific priors influence context-driven fabrication. \textit{Format}: \textcolor{taskMC}{\textbf{MC}} = Multiple Choice, \textcolor{taskBin}{\textbf{Bin}} = Yes/No, \textcolor{taskOpen}{\textbf{Open}} = Open-ended QA, \textcolor{taskCap}{\textbf{Cap}} = Captioning. \textit{Length}: \textcolor{lenS}{\textbf{S}} = Short ($<$1min), \textcolor{lenM}{\textbf{M}} = Medium (1--5min), \textcolor{lenL}{\textbf{L}} = Long ($>$5min), \textcolor{lenSt}{\textbf{St}} = Streaming. \textit{Baseline}: Specialized baseline method proposed. \textit{SOTA Perf.}: Representative best performance reported.}
\label{tab:benchmark_overview}
\resizebox{\textwidth}{!}{%
\begin{tabular}{l l r r l l l l c l}
\toprule
\textbf{Benchmark} & \textbf{Venue} & \textbf{\# Vid} & \textbf{\# QA} & \textbf{Format} & \textbf{Metric} & \textbf{Len} & \textbf{Domain} & \textbf{Baseline} & \textbf{SOTA Perf.} \\
\midrule
\multicolumn{10}{l}{\textit{Spatiotemporal Dynamics Benchmarks (Dynamic Distortion)}} \\
\midrule
VidHalluc~\cite{VidHalluc} & CVPR'25 & 5,002 & 9,295 & \textcolor{taskMC}{\textbf{MC}}, \textcolor{taskBin}{\textbf{Bin}} & Acc, Score & \textcolor{lenS}{\textbf{S}} & ActivityNet, YouCook2, VALOR & \cmark & GPT-4o: 81.2\% \\
VideoHallucer~\cite{VideoHallucer} & arXiv'24 & 948 & 1,800 & \textcolor{taskBin}{\textbf{Bin}} & Acc, Score & \textcolor{lenM}{\textbf{M}} & ActivityNet, VidOR, YouCook & \cmark & Gemini-1.5: 37.8\% \\
HAVEN~\cite{HAVEN} & arXiv'25 & -- & 6.5k & \textcolor{taskMC}{\textbf{MC}}, \textcolor{taskBin}{\textbf{Bin}}, \textcolor{taskOpen}{\textbf{Open}} & Acc, Bias & \textcolor{lenS}{\textbf{S}} & COIN, ActivityNet, Sports1M & \cmark & Valley-Eagle: 61.3\% \\
MHBench~\cite{MHBench} & AAAI'25 & 1,200 & -- & \textcolor{taskMC}{\textbf{MC}}, \textcolor{taskBin}{\textbf{Bin}} & Acc, F1 & \textcolor{lenS}{\textbf{S}} & Sth-Sth V2, Self-shot & \cmark & VideoChat2-MCD: 65.2\% \\
VidHal~\cite{VidHal} & arXiv'24 & 1,000 & 3,000 & \textcolor{taskMC}{\textbf{MC}} & Acc, NDCG & \textcolor{lenS}{\textbf{S}} & TempCompass, MVBench, PT & \xmark & GPT-4o: 77.2\% \\
ARGUS~\cite{ARGUS} & arXiv'25 & 500 & $\sim$9.5k & \textcolor{taskCap}{\textbf{Cap}} & Cost-H/O & \textcolor{lenS}{\textbf{S}} & Ego4D, Panda-70M, Stock & \xmark & Gemini-2.0: 41\% \\
OVBench (THV)~\cite{OVBench} & CVPR'25 & -- & $\sim$33k & \textcolor{taskBin}{\textbf{Bin}} & Accuracy & \textcolor{lenSt}{\textbf{St}} & DiDeMo, QuerYD. & \cmark & VideoChat-On: 63.1\% \\
Vript~\cite{Vript} & NeurIPS'24 & 12k & 420k & \textcolor{taskBin}{\textbf{Bin}}, \textcolor{taskMC}{\textbf{MC}} & Acc, F1 & \textcolor{lenL}{\textbf{L}} & HD-VILA, YouTube, TikTok & \cmark & Vriptor: 58.3 (F1) \\
\midrule
\multicolumn{10}{l}{\textit{Referential Inconsistency Benchmarks (Dynamic Distortion)}} \\
\midrule
EGOILLUSION~\cite{EGOILLUSION} & EMNLP'25 & 1,400 & 8k & \textcolor{taskBin}{\textbf{Bin}}, \textcolor{taskOpen}{\textbf{Open}} & Accuracy & \textcolor{lenS}{\textbf{S}}/\textcolor{lenM}{\textbf{M}} & Ego4D, EgoSeg, Trek-150 & \xmark & Gemini-Pro: 59.4\% \\
MESH~\cite{MESH} & MM'25 & -- & $\sim$140k & \textcolor{taskMC}{\textbf{MC}}, \textcolor{taskBin}{\textbf{Bin}} & Accuracy & \textcolor{lenS}{\textbf{S}}/\textcolor{lenM}{\textbf{M}} & TVQA+, UCF101 & \xmark & GPT-4o: 79.1\% \\
ELV-Halluc~\cite{ELV-Halluc} & arXiv'25 & 200 & 4.8k & \textcolor{taskBin}{\textbf{Bin}} & Acc, SAH & \textcolor{lenL}{\textbf{L}} & YouTube (Event-based) & \cmark & Gemini2.5-Flash: 53.1\% \\
\midrule
\multicolumn{10}{l}{\textit{Context-Driven Fabrication Benchmarks (Content Fabrication)}} \\
\midrule
FactVC~\cite{FactVC} & EMNLP'23 & 300 & -- & \textcolor{taskCap}{\textbf{Cap}} & Bleu4, Rouge-L & \textcolor{lenM}{\textbf{M}}/\textcolor{lenL}{\textbf{L}} & ActivityNet, YouCook2 & \cmark & PDVC-gt: 12.83 (Bleu4) \\
EventHallusion~\cite{EventHallusion} & AAAI'26 & 400 & 711 & \textcolor{taskBin}{\textbf{Bin}}, \textcolor{taskOpen}{\textbf{Open}} & Accuracy & \textcolor{lenS}{\textbf{S}} & ActivityNet & \cmark & GPT-4o: 91.93\% \\
NOAH~\cite{NOAH} & arXiv'25 & 9k & $\sim$60k & \textcolor{taskBin}{\textbf{Bin}}, \textcolor{taskCap}{\textbf{Cap}} & Acc, HR & \textcolor{lenM}{\textbf{M}}/\textcolor{lenL}{\textbf{L}} & ActivityNet & \xmark & Gemini2.5-Flash: 66.8\% \\
VideoHallu~\cite{VideoHallu} & NeurIPS'25 & 987 & 3,233 & \textcolor{taskOpen}{\textbf{Open}} & GPT-Score & \textcolor{lenS}{\textbf{S}} & Generated (Sora, etc.) & \cmark & Comb-GRPO: 57.7 \\
RoadSocial~\cite{RoadSocial} & CVPR'25 & 13.2k & 260k & \textcolor{taskOpen}{\textbf{Open}} & GPT-Score & \textcolor{lenS}{\textbf{S}}/\textcolor{lenM}{\textbf{M}} & Social Media (Traffic) & \xmark & GPT-4o: 69.8  \\
\midrule
\multicolumn{10}{l}{\textit{Audio-Visual Conflict Benchmarks (Content Fabrication)}} \\
\midrule
AVHBench~\cite{AVHBench} & ICLR'25 & 2,136 & 5.3k & \textcolor{taskBin}{\textbf{Bin}} & Accuracy & \textcolor{lenS}{\textbf{S}} & AudioCaps, VALOR & \cmark & AVHModel-Align-FT: 83.9\% \\
CMM~\cite{CMM} & arXiv'24 & 1.2k & 2.4k & \textcolor{taskBin}{\textbf{Bin}} & PA/HR & \textcolor{lenS}{\textbf{S}} & WebVid, AudioCaps & \xmark & Gemini-1.5: 88.4/64.2 \\
EmotionHallucer~\cite{EmotionHallucer} & arXiv'25 & 230 & 2,742 & \textcolor{taskBin}{\textbf{Bin}} & Accuracy & \textcolor{lenS}{\textbf{S}}/\textcolor{lenM}{\textbf{M}} & MER 2023, Social-IQ 2.0 & \cmark & Gemini2.5-Flash: 68.2\% \\
\bottomrule
\end{tabular}%
}
\end{table*}

\section{Evaluation Benchmarks}
\label{sec:benchmarks}

Following the taxonomy in Section~\ref{sec:taxonomy}, we categorize benchmarks by hallucination type and representative failure cases. Table~\ref{tab:benchmark_overview} provides an overview of their venues, scales, task formats, and evaluation metrics, with detailed analysis presented in Appendix~\ref{app:benchmarks}.

\subsection{Dynamic Distortion Benchmarks}
\textbf{Spatiotemporal dynamics benchmarks} assess Vid-LLMs' ability to model temporal structure, covering three subtypes: event misordering, duration distortion, and frequency confusion. 

For \textit{event misordering}, VidHalluc~\cite{VidHalluc} includes 5,002 videos from ActivityNet~\cite{ActivityNet}, YouCook2~\cite{YouCook2}, and VALOR32K~\cite{VALOR32K}, and evaluates temporal hallucinations through sequence-based QA tasks that test whether models can determine the correct order of actions.
HAVEN~\cite{HAVEN} (event) targets discrepancies in action sequences using 2,245 questions across binary, multiple-choice, and short-answer formats.
MHBench~\cite{MHBench} provides 1,200 videos and tests motion understanding via adversarial triplets simulating original, reversed, and incomplete actions.
ARGUS~\cite{ARGUS} evaluates hallucination and omission on 500 videos with about 9,500 annotations, penalizing event misordering by checking the temporal alignment between model-generated and ground-truth action sequences.

For \textit{duration distortion}, VideoHallucer~\cite{VideoHallucer} includes 1,800 adversarial question pairs based on 948 videos, assessing both intrinsic and extrinsic hallucinations through tasks focused on detecting abnormal durations and comparing relative event lengths.
OVBench (THV)~\cite{OVBench} targets duration distortion in real-time streaming settings, requiring models to track action persistence and estimate the length of ongoing events as temporal context unfolds.

For \textit{frequency confusion}, VidHal~\cite{VidHal} benchmarks fine-grained temporal understanding by asking models to distinguish between captions with correct and hallucinated action counts.
HAVEN~\cite{HAVEN} (count) addresses this via numerical questions that test a model's ability to differentiate between single and repeated actions, evaluating its capacity to quantify frequency.
Vript~\cite{Vript} includes a `Count' category in its Vript-RR benchmark, which assesses whether models can accurately compare the number of visual elements across long video sequences.

\textbf{Referential inconsistency benchmarks} assess whether models can maintain distinct representations of entities and scenes over time. Despite the increasing use of Vid-LLMs, only three benchmarks explicitly address this issue.

For \textit{character conflation}, EGOILLUSION~\cite{EGOILLUSION} includes 1,400 egocentric videos and 8,000 question–answer pairs. It evaluates whether models confuse different individuals, for example by identifying the camera wearer as another person during object interactions or activity recognition.
MESH~\cite{MESH} introduces a human-aligned evaluation framework called Mise En Scène, built on TVQA+ clips. It tests whether models can consistently track character identity, appearance, and actions across scenes using structured evaluation traps.

For \textit{scene conflation}, ELV-Halluc~\cite{ELV-Halluc} uses approximately 4,800 adversarial video and text pairs to evaluate whether models incorrectly assign visual elements such as objects or actions from one part of a video to another.

\subsection{Content Fabrication Benchmarks}

\textbf{Context-driven fabrication benchmarks} assess whether models generate outputs based on visual evidence rather than relying on statistical associations from training data. These benchmarks span various domains such as activity recognition, driving, and synthetic videos, reflecting the diverse and context-sensitive nature of fabrication errors.

For \textit{object-action hallucination}, VideoHallu~\cite{VideoHallu} uses synthetic ``negative control'' videos to test whether models incorrectly infer actions based on prior object–action associations instead of actual motion cues. For instance, a model may claim that a watermelon breaks after being shot even when it remains intact in the video.
FactVC~\cite{FactVC} identifies action consistency as a major source of captioning error, accounting for 38.3\% of failures. Models often describe interactions such as a person dancing with a dog based on object co-occurrence, without grounding predictions in visual dynamics.

For \textit{scene-event hallucination}, EventHallusion~\cite{EventHallusion} evaluates whether models hallucinate events by over-relying on typical scene–event pairings, such as assuming cooking takes place in a kitchen even without action evidence.
NOAH~\cite{NOAH} scales this evaluation to over 60,000 samples created from around 9,000 edited videos, testing whether models ignore inserted contradictory clips and instead generate events that align with the surrounding scene or narrative context.
RoadSocial~\cite{RoadSocial} focuses on driving scenarios, using adversarial and incompatible question formats to test whether models hallucinate common road events, such as collisions or traffic violations, based solely on general road context or misleading prompts, even when no such events occur in the video.

\textbf{Audio-visual conflict benchmarks} evaluate whether models integrate audio and visual signals appropriately, focusing on cases where dominant audio cues override visual input and lead to incorrect predictions. With only three existing benchmarks, this category remains underexplored, and current datasets are limited to short video clips. As multimodal Vid-LLMs increasingly process audio, further benchmark development is needed.

For \textit{action attribution}, AVHBench~\cite{AVHBench} tests whether sounds such as music or bird calls cause models to generate incorrect visual descriptions like ``a person is dancing'' or ``a bird is chirping,'' even when no such actions are visible. It includes 2,136 videos and 5,302 binary question–answer pairs sourced from AudioCaps and VALOR, and reports precision, recall, and F1 scores to quantify errors.
CMM~\cite{CMM} evaluates similar cases using curated ``audio dominance'' samples, where prominent sounds such as thunder occur without visual events like lightning. Models are asked binary questions to assess whether they mistakenly rely on audio alone for visual claims.

For \textit{emotion inference}, EmotionHallucer~\cite{EmotionHallucer} examines whether models infer incorrect emotional states based on misleading multimodal cues. Its Reasoning Result and Reasoning Cue tasks test if models describe a neutral face as ``excited'' due to upbeat vocal tone, or invent emotional cues to justify unsupported conclusions.

\subsection{Discussion: Coverage and Gaps}

Table~\ref{tab:benchmark_overview} summarizes 19 existing benchmarks for evaluating video hallucination, with a notable concentration on Spatiotemporal Dynamics (8 benchmarks), mostly targeting short clips. A few, such as Vript~\cite{Vript} and OVBench~\cite{OVBench}, extend to long-form or streaming contexts. Context-Driven Fabrication shows broad domain coverage, ranging from traffic scenarios (RoadSocial) to synthetic videos (VideoHallu). In contrast, Referential Inconsistency and Audio-Visual Conflict remain underexplored, each represented by only three benchmarks, and no benchmark addresses audio-visual consistency in long-form videos. While 11 benchmarks include dedicated baselines to support method development, performance analysis reveals a clear divide: state-of-the-art models perform well on some tasks (e.g., VidHalluc, EventHallusion, with scores above 80\%), but struggle with fine-grained temporal reasoning (e.g., VideoHallucer, 37.8\%) and long-context consistency (e.g., ELV-Halluc, 53.1\%). These findings identify dynamic distortion and long-range temporal grounding as persistent challenges for future research.
\begin{table*}[htbp]
\centering
\caption{Summary of video hallucination mitigation strategies. \textit{Case}: \textcolor{tempEM}{Order}/\textcolor{tempDD}{Dur}/\textcolor{tempFC}{Freq} (Spatiotemporal Dynamics), \textcolor{aggrCA}{Char}/\textcolor{aggrSA}{Scn} (Referential Inconsistency), \textcolor{coocOA}{Obj-Act}/\textcolor{coocSE}{Scn-Evt} (Context-Driven Fabrication), \textcolor{audioAA}{Act}/\textcolor{audioEI}{Emo} (Audio-Visual Conflict). \textit{TF}: \cmark\ = train-free, \xmark\ = training required. \textit{Reported Gain}: reported improvement (over baseline) on primary benchmark.}
\label{tab:mitigation_summary}
\resizebox{\textwidth}{!}{%
\begin{tabular}{l l c c l l l}
\toprule
\textbf{Method} & \textbf{Venue} & \textbf{Case} & \textbf{TF} & \textbf{Core Technique} & \textbf{Key Mechanism} & \textbf{Reported Gain} \\
\midrule
\multicolumn{7}{l}{\textit{Spatiotemporal Dynamics Mitigation (Dynamic Distortion)}} \\
\midrule
SEASON~\cite{SEASON} & arXiv'25 & \textcolor{tempEM}{\textbf{Order}} & \cmark & Contrastive Decoding & Temporal homogenization contrast & +5.7\% Acc (Qwen2.5-VL) \\
Video-thinking~\cite{HAVEN} & arXiv'25 & \textcolor{tempEM}{\textbf{Order}} & \xmark & Preference Optimization & Segment-weighted thinking contrast & +7.4\% Acc (LLaVA-NeXT) \\
SmartSight~\cite{SmartSight} & arXiv'25 & \textcolor{tempEM}{\textbf{Order}} & \cmark & Introspective Sampling & Temporal attention collapse score & +2.9\% Acc (Video-R1) \\
Temporal Insight~\cite{TemporalInsightEnhancement} & ICPR'24 & \textcolor{tempDD}{\textbf{Dur}} & \cmark & Post-hoc Correction & Iconic action timestamp extraction & +27.9\% R@1 (Video-LLaMA) \\
DINO-HEAL~\cite{VidHalluc} & CVPR'25 & \textcolor{tempDD}{\textbf{Dur}} & \cmark & Feature Reweighting & DINOv2 spatial saliency reweighting & +7.0\% Acc (Video-LLaVA) \\
TAAE~\cite{TAAE} & arXiv'25 & \textcolor{tempDD}{\textbf{Dur}} & \xmark & Activation Engineering & Temporal-aware offset injection & +4.4\% Acc (Qwen2.5-VL) \\
VTG-LLM~\cite{VTG-LLM} & AAAI'25 & \textcolor{tempFC}{\textbf{Freq}} & \xmark & Temporal Grounding & Absolute-time token disentanglement & +6.8\% R@1 (Video-LLaMA2) \\
Vriptor~\cite{Vript} & NeurIPS'24 & \textcolor{tempFC}{\textbf{Freq}} & \xmark & Video-Script Alignment & Dense script-based timestamp training & +7.5\% F1 (ST-LLM) \\
\midrule
\multicolumn{7}{l}{\textit{Referential Inconsistency Mitigation (Dynamic Distortion)}} \\
\midrule
Vista-llama~\cite{Vista-llama} & CVPR'24 & \textcolor{aggrCA}{\textbf{Char}} & \xmark & Token Processing & Equal distance visual attention & $\sim$5.0\% Acc (LLaVA) \\
VideoPLR~\cite{VideoPLR} & arXiv'25 & \textcolor{aggrCA}{\textbf{Char}} & \xmark & Perception-Logic-Reasoning & Database-anchored symbolic execution & +9.2\% Acc (Qwen2.5-VL) \\
ELV-Halluc-DPO~\cite{ELV-Halluc} & arXiv'25 & \textcolor{aggrSA}{\textbf{Scn}} & \xmark & Preference Optimization & Cross-segment adversarial DPO & -27.7\% SAH (Qwen2.5-VL) \\
VideoChat-Online~\cite{OVBench} & CVPR'25 & \textcolor{aggrSA}{\textbf{Scn}} & \xmark & Streaming Processing & Pyramid memory bank update & +8.5\% Acc (InternVL2)\\
\midrule
\multicolumn{7}{l}{\textit{Context-Driven Fabrication Mitigation (Content Fabrication)}} \\
\midrule
SANTA~\cite{SANTA} & arXiv'25 & \textcolor{coocOA}{\textbf{Obj-Act}} & \xmark & Fine-grained Contrastive Tuning & Hard negative action/object swapping & +2.4\% Acc (LLaVA-Video) \\
TCD~\cite{EventHallusion} & AAAI'26 & \textcolor{coocOA}{\textbf{Obj-Act}} & \cmark & Contrastive Decoding & Logit subtraction of priors & +3.2\% Acc (VILA) \\
MASH-VLM~\cite{MASH-VLM} & CVPR'25 & \textcolor{coocSE}{\textbf{Scn-Evt}} & \xmark & Disentangled Representation & DST-Attention \& Harmonic-RoPE & +2.7\% Acc (ST-LLM) \\
PaMi-VDPO~\cite{PaMi-VDPO} & arXiv'25 & \textcolor{coocSE}{\textbf{Scn-Evt}} & \xmark & Preference Optimization & Part-mismatch visual negatives & +5.9\% Acc (LLaVA-OenVision) \\
MMA~\cite{MMA} & IJCAI'25 & \textcolor{coocSE}{\textbf{Scn-Evt}} & \xmark & Parameter-Efficient Tuning & Dual-path visual-text alignment & +2.0\% Acc (MA-LMM) \\
VistaDPO~\cite{VistaDPO} & ICML'25 & \textcolor{coocOA}{\textbf{O-A}}, \textcolor{coocSE}{\textbf{S-E}} & \xmark & Visual-State DPO & Penalizing low visual-dependency tokens & +36.5\% Acc (Video-LLaVA)\\
VideoHallu-GRPO~\cite{VideoHallu} & NeurIPS'25 & \textcolor{coocOA}{\textbf{O-A}}, \textcolor{coocSE}{\textbf{S-E}} & \xmark & RL Fine-Tuning & Group-relative rewards on counter-intuitive data & +4.7\% Acc (Qwen2.5-VL)\\
\midrule
\multicolumn{7}{l}{\textit{Audio-Visual Conflict Mitigation (Content Fabrication)}} \\
\midrule
AVHModel-Align-FT~\cite{AVHBench} & ICLR'25 & \textcolor{audioAA}{\textbf{Act}} & \xmark & Instruction Tuning & Modality-Disentangled Data & +33.8\% Acc (Video-LLaMA)\\
AVCD~\cite{AVCD} & NeurIPS'25 & \textcolor{audioAA}{\textbf{Act}} & \cmark & Trimodal Contrastive Decoding & Dominance-aware Attentive Masking & +1.6\% Acc (VideoLLaMA2) \\
PEP-MEK~\cite{EmotionHallucer} & arXiv'25 & \textcolor{audioEI}{\textbf{Emo}} & \cmark & Predict-Explain-Predict & Knowledge Extraction \& Refinement & +9.1\% Acc (Gemini2.5-Flash) \\
\bottomrule
\end{tabular}%
}
\end{table*}
\section{Mitigation Strategies}
\label{sec:mitigation}

Following the taxonomy in Section~\ref{sec:taxonomy}, we group mitigation strategies by hallucination type and representative failure cases. Table~\ref{tab:mitigation_summary} summarizes the corresponding techniques, with detailed analysis provided in Appendix~\ref{app:methods}.

\subsection{Mitigating Dynamic Distortion}

\textbf{Spatiotemporal dynamics mitigation} tackles event order, duration, and frequency Hallucinations using contrastive, optimization-based, and temporal grounding strategies, with event misordering being the most extensively studied.

For \textit{event misordering}, SEASON~\cite{SEASON} contrasts original videos with temporally homogenized negatives that disrupt causal order, using self-diagnostic decoding to suppress outputs insensitive to correct sequence. Video-thinking~\cite{HAVEN} introduces TDPO (Thinking-based DPO), applying segment-weighted preference learning on reasoning paths to optimize for temporal logic. SmartSight~\cite{SmartSight} ranks multiple responses based on the Temporal Attention Collapse (TAC) score, favoring outputs that attend proportionally across time to preserve correct order.

For \textit{duration distortion}, Temporal Insight Enhancement~\cite{TemporalInsightEnhancement} decomposes events into atomic actions and leverages external vision models to timestamp them, aligning model responses with grounded temporal claims. DINO-HEAL~\cite{VidHalluc} uses DINOv2-guided spatial saliency to reweight features and maintain attention on action-relevant regions across time. TAAE~\cite{TAAE} identifies activation offsets between full and downsampled inputs to amplify duration-sensitive representations during inference.

For \textit{frequency confusion}, VTG-LLM~\cite{VTG-LLM} introduces absolute-time tokens that decouple event identity from repetition count, improving temporal anchoring of repeated actions. Vriptor~\cite{Vript} aligns dense scene-level captions with timestamps to enforce instance-level discrimination, helping models avoid merging or duplicating repeated events in long videos.

\textbf{Referential inconsistency mitigation} focuses on preserving distinct representations of entities and scenes across time.

For \textit{character conflation}, Vista-LLaMA~\cite{Vista-llama} introduces Equal Distance Attention, which removes positional decay between visual and textual tokens, ensuring stable attention to character identities regardless of when they appear. VideoPLR~\cite{VideoPLR} constructs a structured video database with explicit object tracking, allowing symbolic logic programs to differentiate entity identities during reasoning.

For \textit{scene conflation}, ELV-Halluc-DPO~\cite{ELV-Halluc} applies adversarial preference optimization using cross-segment perturbations that swap entities in space or time, encouraging the model to ground predictions within the correct segment. VideoChat-Online~\cite{OVBench} uses a Pyramid Memory Bank during streaming inference to separate recent high-resolution frames from compressed long-term history, reducing confusion between distinct temporal contexts.

\subsection{Mitigating Content Fabrication}

\textbf{Context-driven fabrication mitigation} aims to separate predictions from statistical associations and strengthen visual grounding.

For \textit{object–action hallucination}, methods emphasize motion cues over object presence. SANTA~\cite{SANTA} applies fine-grained contrastive tuning using hard negatives where actions differ but entities remain the same, encouraging the model to rely on motion rather than co-occurrence patterns. TCD~\cite{EventHallusion} suppresses object-triggered predictions by subtracting logits from temporally shuffled inputs, guiding the model to attend to dynamic cues rather than static priors.

For \textit{scene–event hallucination}, methods reduce the influence of background context on event inference. MASH-VLM~\cite{MASH-VLM} uses disentangled spatial-temporal attention to prevent the model from predicting actions based solely on static backgrounds. PaMi-VDPO~\cite{PaMi-VDPO} introduces preference learning with visually mismatched negatives to train the model to verify event descriptions against the actual scene. MMA~\cite{MMA} aligns local visual details with textual tokens through a dual-path adapter, reinforcing grounding in specific cues over general context.

Some methods address both cases. Vista-DPO~\cite{VistaDPO} penalizes predictions that lack visual grounding by optimizing against prior-driven outputs, encouraging reliance on directly observable evidence. VideoHallu-GRPO~\cite{VideoHallu} improves grounding using synthetic videos with counterintuitive scenarios, optimizing model behavior through group-based relative rewards that favor visually consistent responses over learned priors.

\textbf{Audio-visual conflict mitigation} addresses errors caused by dominant audio signals overriding visual input. This area remains underexplored, with few targeted methods.

For \textit{action attribution}, AVHModel-Align-FT~\cite{AVHBench} fine-tunes on annotations separating audio and visual events, helping models distinguish between auditory and visual sources. AVCD~\cite{AVCD} uses contrastive learning with modality masking to suppress misleading cues and improve cross-modal grounding.

For \textit{emotion inference}, PEP-MEK~\cite{EmotionHallucer} enforces modality-specific reasoning by requiring models to explain visual evidence before integrating it with audio, reducing overreliance on vocal tone.

\subsection{Discussion: Coverage and Trade-offs}
Table~\ref{tab:mitigation_summary} reveals uneven progress across hallucination types. While Spatiotemporal Dynamics and Context-Driven Fabrication are well addressed, Referential Inconsistency and Audio-Visual Conflict remain underexplored. A key trade-off exists between effectiveness and deployment cost. Training-based methods (e.g., VistaDPO~\cite{VistaDPO}, AVHModel-Align-FT~\cite{AVHBench}) offer substantial gains (up to 30–36\%) by reshaping model priors via reinforcement learning or instruction tuning, but require high training overhead. In contrast, training-free strategies (e.g., SEASON~\cite{SEASON}, SmartSight~\cite{SmartSight}, TCD~\cite{EventHallusion}) are model-agnostic and easier to adopt, though with more limited gains and occasionally higher inference latency. Mechanism suitability often aligns with error type: inference-time decoding or attention adjustments help correct temporal and logical inconsistencies (e.g., VideoPLR~\cite{VideoPLR}), while hallucinations driven by strong priors call for deeper disentanglement during training (e.g., MASH-VLM~\cite{MASH-VLM}). Reducing the latency of current inference-time methods is critical for real-time and safety-sensitive deployment.
\section{Future Directions}
\label{sec:future}

Building on the taxonomy in Section~\ref{sec:taxonomy}, we propose two core directions for enhancing hallucination robustness in Vid-LLMs, targeting the underlying causes of dynamic distortion and content fabrication.

\subsection{Addressing Dynamic Distortion: Temporal and Referential Fidelity}

Dynamic distortion primarily results from a gap between visual encoding and temporal reasoning. This issue is often rooted in the use of static image encoders and pooling-based connectors, which tend to discard motion cues critical for capturing temporal dynamics~\cite{EncNoDynamic,VideoChat,Video-LLaMA}. As videos become longer, this problem is further exacerbated by the limited capacity of current models to maintain long-range context, leading to semantic drift and referential inconsistency~\cite{StreamingLLM,VTG-LLM}.

To address these limitations, future architectures should adopt video-oriented designs that preserve temporal structure throughout the pipeline. This includes using video-native encoders such as VideoMAE~\cite{VideoMAEv2} and motion-aware connectors that incorporate signals like optical flow~\cite{Flow4agent} to capture velocity and trajectory. For long-range consistency, models may benefit from structured memory mechanisms, including state space models like Mamba~\cite{VideoMamba} or episodic memory ~\cite{VideoEM}, to retain persistent entity information over extended sequences. 

\subsection{Mitigating Content Fabrication: Grounding and Alignment}

Content fabrication arises when pretraining priors dominate over visual grounding. Models may hallucinate actions or events based on static entities or scenes, ignoring temporal evidence~\cite{SANTA,MASH-VLM}. The problem is worsened by imbalanced modality integration, where dominant audio cues override visual input, leading to cross-modal conflicts~\cite{AVHBench,CMM}. 

To reduce fabrication, models should learn to separate priors from perceptual evidence. This can be achieved by using counterfactual strategies, such as introducing negative samples with implausible object–action pairs and applying debiasing objectives that encourage reliance on motion cues~\cite{BSSARD}. In audio-visual settings, models should verify visual input before incorporating audio signals to avoid hallucinations caused by sound~\cite{AVAgent}.

\section{Generalization to Emerging Settings}
While this survey focuses on core Vid-LLM capabilities, the proposed taxonomy generalizes to emerging settings by grounding categories in observable output manifestations rather than model architectures or task formats. We illustrate potential risks and their mappings in representative settings.

\textbf{Very long videos.} Exacerbate cross-segment inconsistency and long-range temporal errors. Distortions in order, duration, and frequency become more prevalent, while character or scene drift across distant segments maps to long-range Dynamic Distortion.

\textbf{Interactive or streaming settings.} Introduce errors under incomplete or evolving evidence, including premature conclusions and failure to update predictions. Visually supported but mislocalized or misordered events are classified as Dynamic Distortion, while unsupported or audio-dominant claims fall under Content Fabrication. Classification is applied per claim within its temporal window.

\textbf{Agentic Vid-LLMs.} Introduce upstream risks such as incorrect retrieval, memory contamination, or over-reliance on tools. These manifest as relation mis-modeling despite visual support (Dynamic Distortion) or unsupported assertions under insufficient evidence (Content Fabrication).

\section{Conclusion}
\label{sec:conclusion}

Video large language models (Vid-LLMs) have achieved significant progress in video-language modeling, but also give rise to unique hallucination patterns that differ from those in static image tasks. 
This survey introduces a mechanism-based taxonomy that categorizes hallucinations in Vid-LLMs into two primary categories: Dynamic Distortion, referring to the misinterpretation of spatiotemporal progression or referential consistency; and Content Fabrication, referring to ungrounded outputs influenced by statistical context priors or dominant auditory cues. 
Although recent studies have advanced benchmarking and mitigation of spatiotemporal and context-driven hallucinations, challenges such as referential inconsistency and audio-visual conflicts remain underexplored. 
Furthermore, most existing mitigation strategies are applied at inference time or as post-training adjustments, highlighting the need for scalable, training-time alignment methods. 
To improve the robustness of Vid-LLMs, we advocate future research toward developing video-native encoders that preserve motion cues, integrating explicit memory mechanisms to support long-term temporal grounding, and employing counterfactual learning to disentangle model reasoning from prior-driven associations. These directions will be essential for building trustworthy and temporally faithful video-language systems.

\section*{Limitations}
Previous surveys on hallucination in MLLMs have extended the scope from LLMs and image-based VLMs to include video and other modalities. However, their coverage of video hallucination remains limited, with only brief mentions of benchmarks and mitigation efforts, lacking structured categorization or causal analysis. This survey addresses this gap by presenting a mechanism-driven taxonomy of hallucinations in Vid-LLMs. We introduce a layered classification framework, review recent studies in greater depth, analyze the underlying causes of hallucinations, and outline future research directions. While we have strived to cover key developments in Vid-LLM hallucination, some relevant work may be omitted. This survey includes research published up to January 2026.

\section*{Ethics Statement}
This survey adheres to established ethical standards for academic research. All referenced works are publicly available, and no human subjects or personally identifiable information are involved. The purpose of this survey is to facilitate academic understanding and encourage responsible development in the study of hallucination in Vid-LLMs. All prior work has been properly cited, with appropriate credit given to original contributions.

\bibliography{custom}

\appendix

\section{Detailed Analysis of Benchmarks}
\label{app:benchmarks}

Benchmarks for video hallucination differ in format, domain coverage, and temporal scale, each introducing distinct strengths and limitations.

\textbf{Generative vs. Discriminative formats.} Generative benchmarks (e.g., ARGUS, VideoHallu, RoadSocial) based on open-ended QA or captioning provide deeper insights into free-form reasoning, but suffer from evaluation bottlenecks, often relying on costly LLM-as-a-judge scoring that may introduce bias or secondary hallucinations. In contrast, discriminative benchmarks (e.g., VidHalluc, VideoHallucer, MHBench) using multiple-choice or binary QA offer standardized and scalable metrics (e.g., Accuracy, F1), yet may encourage shortcut learning via linguistic artifacts rather than true visual grounding.

\textbf{Domain-specific vs. General-purpose.} General-purpose benchmarks (e.g., Vript, NOAH) span diverse domains (e.g., ActivityNet, TikTok, YouTube), enabling broad capability assessment. Domain-specific benchmarks, however, are critical for specialized applications: RoadSocial emphasizes safety-critical traffic scenarios with dynamic distortions, while AVHBench focuses on audio-visual conflict.

\textbf{Temporal Scale.} Long-video and streaming benchmarks (e.g., OVBench, ELV-Halluc) stress-test referential consistency and long-range temporal reasoning, making them essential for evaluating production-level systems. In contrast, short-clip benchmarks (e.g., HAVEN, CMM) better isolate fine-grained action perception and cross-modal alignment without long-range memory effects.

\section{Detailed Analysis of Methods}
\label{app:methods}

Mitigation strategies can be compared in terms of effectiveness, efficiency, and deployment constraints, revealing distinct trade-offs across different methodological paradigms.

\textbf{Contrastive decoding and inference-time interventions} (e.g., SEASON, SmartSight, TCD, TAAE) are model-agnostic and training-free, enabling flexible deployment without additional data collection or retraining. However, these approaches typically incur increased inference latency due to multiple forward passes and rely on heuristic sampling or attention manipulation, which may limit their effectiveness in overcoming strong parametric priors. As a result, they are most suitable for rapid deployment scenarios where retraining is infeasible and errors are primarily local or temporal.

\textbf{Supervised fine-tuning and alignment} (e.g., AVHModel-Align-FT, Vriptor, VTG-LLM) directly improve representation quality, particularly for multimodal alignment and disentangling spurious correlations. Their effectiveness, however, is constrained by the availability of high-quality annotated data and the substantial cost of retraining. These methods are therefore best suited for domain-specific settings where sufficient data and computational resources are available.

\textbf{Preference optimization and reinforcement learning} (e.g., HAVEN, ELV-Halluc-DPO, VistaDPO) achieve strong performance gains by explicitly reshaping model priors and penalizing hallucination-prone behaviors. This performance comes at the cost of high computational complexity, including adversarial data generation and reward model design. Such approaches are most appropriate for late-stage alignment of high-performance Vid-LLMs, where robustness under diverse conditions is critical.

\textbf{Architectural and symbolic modifications} (e.g., Vista-LLaMA, MASH-VLM, VideoPLR, Temporal Insight) address fundamental limitations such as positional decay, static bias, and weak structural grounding, while symbolic components additionally provide interpretability. These benefits are offset by high implementation costs, including the need for pretraining from scratch or increased system complexity and latency. Consequently, these methods are well suited for next-generation model design or safety-critical applications requiring strong interpretability and reliable grounding.

\end{document}